\documentclass{llncs}
\usepackage{amsmath}
\usepackage{amssymb}
\usepackage{color}
\usepackage[linesnumbered,longend,ruled,vlined]{algorithm2e}
\usepackage{tikz}
\usepackage{pgfplots}
\usepackage{booktabs}
\usetikzlibrary{arrows}
\usepackage{subcaption}
\usepackage{multicol}
\usepackage{changes}
\usepackage{rotating}
\usepackage{longtable}
\usepackage{url}
\usepackage{hyperref}
\pgfplotsset{compat=1.16} 

\begin{document}
\title{Exploitation Strategies in Conditional Markov Chain Search: A case study on the three-index assignment problem}
%
%
\author{Sahil Patel\inst{1} \and Daniel Karapetyan\inst{1}}
\institute{Computer Science, University of Nottingham, Nottingham NG8\,1BB, UK, \email{sahil.j.patel@gmail.com}, \email{daniel.karapetyan@nottingham.ac.uk}}
%
%
%
\maketitle

\begin{abstract}
The Conditional Markov Chain Search (CMCS) is a framework for automated design of metaheuristics for discrete combinatorial optimisation problems.
Given a set of algorithmic components such as hill climbers and mutations, CMCS decides in which order to apply those components.
The decisions are dictated by the CMCS configuration that can be learnt offline. CMCS does not have an acceptance criterion; any moves are accepted by the framework. As a result, it is particularly good in exploration but is not as good at exploitation. 
In this study, we explore several extensions of the framework to improve its exploitation abilities.
To perform a computational study, we applied the framework to the three-index assignment problem.
The results of our experiments showed that a two-stage CMCS is indeed superior to a single-stage CMCS.

\end{abstract}

\section{Introduction}
\label{sec:cmcs}
\newcommand{\Mfail}{M^\text{fail}}
\newcommand{\Msucc}{M^\text{succ}}
\newcommand{\ComponentPool}{\mathcal{H}}
The Conditional Markov Chain Search (CMCS) is a framework based on a highly-configurable meta-heuristic suitable for automated design of combinatorial optimisation algorithms.
We will say that a \emph{CMCS configuration} is a specific metaheuristic, as opposed to the generic CMCS framework.
A CMCS configuration comprises a set of components such as hill climbers and mutations, and a control mechanism that decides in which order those components are applied.
To build a CMCS configuration, the user needs to provide a component pool, an objective function and a set of test instances.
Using this information, a \emph{configurator} searches for a high-performance CMCS configuration.

A CMCS configuration is a single-point meta-heuristic based on multiple components treated as black boxes. 
Each component is a subroutine that takes a solution and modifies it according to the internal logic. 
Some components such as hill climbers may aim at improving solutions whereas other components such as mutations may perform random modifications; the framework generally treats them the same. 
The control mechanism of a CMCS configuration is described by a set of numeric parameters thus enabling automated generation of CMCS configurations; by tuning these parameters, one can find the `optimal' control mechanism~\cite{Karapetyan2018}.

A CMCS configuration performs as follows. 
It takes as an input the initial solution (usually produced by some construction heuristic, e.g. random solution) and then applies to it one of the components at each iteration. 
Any modification by any component is always `accepted', i.e. there is no backtracking. 
CMCS records whether the component improved the solution or not. 
The choice of the next component depends only on which component was used in the current iteration and whether it improved the solution; thus the sequence of applied components is a Markov chain.
The control mechanism can be defined by two transition matrices: one for the case when the solution was improved ($\Msucc$) and another one for the case when the solution was not improved ($\Mfail$). Transitions may be deterministic (if there is exactly one non-zero value in each row of each matrix) or probabilistic. 
Since CMCS may worsen the solution, it keeps track of the best solution found during the search and at the end returns that solution. 
The termination criterion is based on the user-defined time budget~\cite{Karapetyan2018}.

An important aspect in an optimisation heuristic is how it combines exploration and exploitation. 
By \emph{exploration} we mean the ability to find diverse solutions and by \emph{exploitation} we mean the ability to improve upon found solutions.
If a heuristic does not do enough exploration, it converges prematurely.
If it does not do enough exploitation, it misses good solutions even if it succeeds in finding the regions where they are.
By configuring CMCS, one can achieve an optimal balance between exploration and exploitation, however this balance will mainly be static, as CMCS is trained offline and has limited online adaptiveness capabilities.
In this paper, we study if adjusting the balance between exploration and exploitation dynamically could benefit the performance of CMCS configurations.

Specifically, we propose and study two extensions of CMCS that enable dynamic adjustment of the exploration/exploitation balance.
Note that the original CMCS does not have any mechanisms that would allow adaptation based on the quality of the current solution.
We explore two approaches to implement such adaptation: one triggered by finding a new best-found-so-far solution (such solutions, obviously, tend to be good), and another one based on the elapsed time, as the solution quality tends to improve throughout the solution process.

As a case study, we use the Three-Index Assignment Problem (AP3)\@. 
AP3 is the three-dimensional extension of the job Assignment Problem.
While the Assignment Problem is known to by polynomially solvable, AP3 is NP-hard.

Applications of AP3 include military troop assignment, minimising idle time in a rolling mill, scheduling capital investments, satellite coverage and cost optimisation~\cite{Pierskalla}, production of printed circuit boards~\cite{Crama1992}, and scheduling a teaching practice~\cite{Frieze}.

The contributions of the paper are as follows.  
In all previous applications of CMCS~\cite{Karapetyan2017,Karapetyan2018,Karapetyan2019}, only one strategy of exploration and exploitation has been applied. 
Hence, to address this gap, we propose two new strategies and compare all three strategies in a computational study.
To do so, we also design a CMCS configurator and a pool of AP3 components.

The remainder of this paper is structured as follows. 
In Section~\ref{sec:cmcs-strategies}, the different CMCS strategies are detailed, followed by the CMCS configurator (Section~\ref{sec:configurator}), discussion of AP3 and the component pool (Sections~\ref{sec:ap3} and~\ref{sec:ap3-components}, respectively), experimental evaluation (Section~\ref{sec:evaluation}) and conclusions and future work (Section~\ref{sec:conclusions}).

\section{CMCS Exploitation Strategies}
\label{sec:cmcs-strategies}

As we mentioned earlier, in each iteration, CMCS selects one component and applies it to the current solution.
The selection of the component depends on exactly two inputs: what was the previously applied component and whether it improved the solution.
This means that the control mechanism cannot respond to the absolute quality of the current solution; it only `knows' if the solution was improved or not in the last iteration.
As a result, it cannot adjust its exploitation strength when a particularly good solution is found.
In other words, when it finds a good solution, it may keep exploring without an attempt to improve this solution.

In this section, we describe three versions of CMCS that address the exploitation in different ways.
Strategy~A stands for the original CMCS, strategy~B stands for a new extension that changes its behaviour when a new best-found-so-far solution is obtained, to intensify exploitation of particularly good solutions.
Strategy~C stands for a new extension that splits the time budget into two phases, where the first phase is expected to be focused on exploration while the second phase is focused on exploitation.

\subsection{$\text{CMCS}^\text{A}$: Strategy A}
Strategy~A is the core CMCS algorithm~\cite{Karapetyan2018}.
Its behaviour makes it very good at exploration of the solution space but it may not fully exploit good solutions before proceeding to further exploration.

The details of Strategy~A are shown in Algorithm~\ref{alg:cmcs-a}.
It takes the following as the input:
\begin{itemize}
    \item 
    Ordered set $\ComponentPool$ of components, called \emph{component pool}.
    Each component is an algorithm that takes a solution and the instance data as parameters and returns an updated solution.

    \item
    Transition matrices $\Msucc$ and $\Mfail$ of size $|\ComponentPool| \times |\ComponentPool|$.
    Each row of each of $\Msucc$ and $\Mfail$ adds up to 1, as these are transition probabilities.

    \item
    Instance data $\mathcal{I}$.
    The instance data is only used for calculating the objective function.

    \item
    Objective function $f(S, \mathcal{I})$, where $S$ is the solution and $\mathcal{I}$ is the instance data.

    \item
    Time budget $T$.
\end{itemize}
The output of the algorithm is the best solution found during the search.

Here $\mathit{RouletteWheel}(p_1, p_2, \ldots, p_n)$ is a function that returns integer $i$ between 1 and $n$ with probability $p_i$.

\begin{algorithm}[tb]
    
    $S^* \gets S_0$; $S \gets S_0$; $f^* \gets f(S_0, \mathcal{I})$; $f_\text{prev} \gets f^*$; $h \gets 1$\;
    \While {$\mathit{elapsed\text{-}time} < T$}
    {
        $S \gets \ComponentPool_h(S, \mathcal{I})$\;
        $f_\text{cur} \gets f(S, \mathcal{I})$\;
        \If {$f_\text{cur} < f_\text{prev}$}
        {
    	    $h \gets \mathit{RouletteWheel}(\Msucc_{h, 1}, \Msucc_{h, 2}, \ldots, \Msucc_{h, |\ComponentPool|})$\;
            
            \If {$f_\text{cur} < f^*$}
            {
                $S^* \gets S$\;
                $f^* \gets f_\text{cur}$\;
            }
        }
    	\Else
    	{
    	    $h \gets \mathit{RouletteWheel}(\Mfail_{h, 1}, \Mfail_{h, 2}, \ldots, \Mfail_{h, |\ComponentPool|})$\;
        }
        
        $f_\text{prev} \gets f_\text{cur}$\;
    }
    
    \Return {$S^*$}\;

\caption{$\text{CMCS}^\text{A}(S, T)$: Strategy A}
\label{alg:cmcs-a}
\end{algorithm}

\subsection{$\text{CMCS}^\text{B}$: Strategy B}

Strategy B is an extension of Strategy A\@.
In order to actively exploit good solutions, it incorporates the variable neighbourhood descent (VND) algorithm (a deterministic local search heuristic that explores a number of neighbourhood structures~\cite{Duarte2018})
that consists of a subset of the hill climbers from the component pool.
Given an ordered set of hill-climbers $\{ \mathcal{HC}_1, \mathcal{HC}_2, \ldots, \mathcal{HC}_n \}$, VND applies $\mathcal{HC}_1$ to the solution as long as it improves the solution.
If $\mathcal{HC}_1$ fails to improve the solution, $\mathcal{HC}_2$ is applied.
If $\mathcal{HC}_2$ improves the solution, VND gets back to $\mathcal{HC}_1$; otherwise it proceeds to $\mathcal{HC}_3$, etc.
If all the hill climbers fail to improve the solution, VND terminates.
For details, see Algorithm~\ref{alg:vnd}.

In essence, Strategy B sacrifices some time to apply the VND algorithm to the best-found-so-far solution each time the algorithm finds a new best-found-so-far solution.
As a result, best-found-so-far solutions are guaranteed to be local minima with respect to the hill climbers included in the VND\@.

Typically, a metaheuristic improves the best-found-so-far solutions particularly frequently at the beginning of its run.
Applying VND each time is unnecessary; it is unlikely that a solution found at the early stages of the search will be better than solutions found later.
Thus, we do not apply VND until $0.5 T$.
For details, see Algorithm~\ref{alg:cmcs-b}.


\begin{algorithm}[tb]

$S^* \gets S_0$; $S \gets S_0$; $f^* \gets f(S_0, \mathcal{I})$; $f_\text{prev} \gets f^*$; $h \gets 1$\;
$f_\text{polished}^*  \gets \infty$; $f_\text{best}^* \gets \infty$\;

$\mathit{vnd\text{-}applied} \gets 0 $\;
\While {$\mathit{elapsed\text{-}time} < T$}
{
    $S \gets \ComponentPool_h(S, \mathcal{I})$\;
    $f_\text{cur} \gets f(S, \mathcal{I})$\;
    \If {$f_\text{cur} < f_\text{prev}$}
    {
	    $h \gets \mathit{RouletteWheel}(\Msucc_{h, 1}, \Msucc_{h, 2}, \ldots, \Msucc_{h, |\ComponentPool|})$\;
        
        \If {$f_\text{cur} < f^*$}
        {
            $S^* \gets S$\;
            $f^* \gets f_\text{cur}$\;
            $\mathit{vnd\text{-}applied} \gets 0$\;
        }
    }
    \Else
    {
	    $h \gets \mathit{RouletteWheel}(\Mfail_{h, 1}, \Mfail_{h, 2}, \ldots, \Mfail_{h, |\ComponentPool|})$\;
    }
    \If {$\mathit{elapsed\text{-}time} \ge 0.5 T$ and $\mathit{vnd\text{-}applied} = 0$}
    {
        $S_\text{polished}^* \gets \text{VND}(S^*)$\;
        $f_\text{polished}^* \gets f(S_\text{polished}^*)$\;
        \If {$f_\text{polished}^*< f_\text{best}^*$}
        {
            $S_\text{best}^* \gets S_\text{polished}^*$\;
        }
        $\mathit{vnd\text{-}applied} \gets 1 $\;
    }
    $f_\text{prev} \gets f_\text{cur}$\;
}

\Return {$S_{best}^*$}\;

\caption{$\text{CMCS}^\text{B}$: Strategy B}
\label{alg:cmcs-b}
\end{algorithm}

\begin{algorithm}[tb]

$ i \gets 1 $\;
\While {$\mathit{elapsed\text{-}time} < T$}
{
    $S_\text{new} \gets \mathcal{HC}_i(S^*, \mathcal{I})$\;
    $f_\text{new} \gets f(S_\text{new}, \mathcal{I})$\;
    \If {$f_\text{new} < f^*$}
    {
        $S^* \gets S_\text{new}$\;
        $f^* \gets f_\text{new}$\;
        $i \gets 1$\;
    }
    \Else
    {
        \If {$i = n$}
        {
            \Return {$S^*$}\;
        }

        $i \gets i + 1$\;      
    }
}
\Return {$S^*$}\;

\caption{Variable Neighbourhood Descent: $\text{VND}(S^*)$}
\label{alg:vnd}
\end{algorithm}


\subsection{$\text{CMCS}^\text{C}$: Strategy C}

Strategy~C splits the time budget into two phases, where the first phase is intended mainly for exploration whereas the second phase is intended mainly for exploitation.
Both phases are implemented as $\text{CMCS}^\text{A}$ although the configurations can be different; we call them `sub-configuration 1' and `sub-configuration 2', respectively.
For details see Algorithm~\ref{alg:cmcs-c}.

Please note that the roles of sub-configuration~1 and~2 are arbitrary; it is up to the configurator to choose what sub-configurations work best within $\text{CMCS}^\text{C}$.
It is our expectation that sub-configuration~1 is more likely to prioritise exploration whereas sub-configuration~2 is more likely to prioritise exploitation.

\begin{algorithm}[tb]


$S^* \gets \text{CMCS}^{A}(S_0, 0.8 T)$ executed with configuration 1\;
$S^* \gets \text{CMCS}^{A}(S^*, 0.2 T)$ executed with configuration 2\;

\Return {$S^*$}\;

\caption{$\text{CMCS}^{C}(S, T)$: Strategy C}
\label{alg:cmcs-c}
\end{algorithm}

\section{CMCS Configurator}
\label{sec:configurator}

This section details the CMCS configurator, i.e.\ the algorithm that produces a CMCS configuration given a training set of problem instances and a component pool.

Each variation of CMCS requires a slightly different configurator but the core algorithm is the same.
The configurator for $\text{CMCS}^\text{A}$ performs as follows. 
It takes as input a time budget $T$ of a single CMCS run, a training set of problem instances and a component pool.
It also takes the number of components that should be included in the configuration.
It then generates all `meaningful' subsets of components of the given size.
A subset is meaningful if it includes at least one hill climber and at least one mutation.

Then, for each meaningful subset of components, the configurator searches for a good combination of the transition matrices.
For that, it employs a simple population-based algorithm that performs as follows:
\begin{enumerate}
    \item
    Produce a population of 50 configurations with the given set of components and random deterministic transition matrices.
    A transition matrix is deterministic if every row contains only one non-zero element.

    \item
    Choose the best configuration in the population.
    If it is better than the best-observed-so-far configuration observed so far, update the best-observed-so-far configuration.

    \item
    Produce a new population: 25 `children' of the best configuration in the previous population and 25 `children' of the best-observed-so-far configuration.
    Each child configuration is obtained from the parent configuration by applying a mutation.
    The configuration mutations are described in Section~\ref{sec:configuration-mutations}.

    \item
    Repeat steps 2--4 for as long as the training time budget allows.

    \item
    Return the best-observed-so-far configuration.
\end{enumerate}

Each evaluation of a configuration consists of running CMCS with the given configuration on a set of benchmark instances.
The average objective value is used as the measure of the configuration quality.


Once optimised configurations are produced for each meaningful subset of components, each of them is tested on an evaluation benchmark set; the one with the highest score is returned.







\bigskip

To configure $\text{CMCS}^\text{B}$, we employ exactly the same algorithm as above.
The VND is predetermined by the user so only the set of components and the transition matrices need to be optimised.

\bigskip

A configuration of CMCS$^C$ consists of two independent CMCS sub-configu\-rations.
Hence, we need to choose two meaningful subsets of components: one for sub-configuration~1 and another one for sub-configuration~2.
For each combination, we then optimise the two pairs of transition matrices.
Specifically, we run the above algorithm to optimise the transition matrices for sub-configuration~1 with sub-configuration~2 disabled.
Then we fix the matrices in sub-configuration~1 and optimise sub-configuration~2 matrices.
Then we fix sub-configuration~2 matrices and optimise again sub-configuration~1 matrices.
Once optimised matrices are produced for each pair of meaningful subsets of components, each of them is tested on an evaluation benchmark set and the one with the highest score is returned.

\subsection{Configuration mutations}
\label{sec:configuration-mutations}

To produce a `child' configuration from a `parent' configuration, the algorithm selects two mutation operators and then applies the first one to $M^\text{succ}$ and the second one to $M^\text{fail}$.
The mutation operators are chosen uniformly at random from a set of available operators, and the choices for the two matrices are completely independent.

Below we describe the configuration matrix mutation operators.

\paragraph{Swap Rows.}
    This mutation uniformly at random chooses two rows of the matrix and swaps them.

\paragraph{Shuffle Row.}
    This mutation uniformly at random chooses a row in the matrix and a number $n$, $0 \le n \le |\mathcal{H}|$.
    It proceeds by making $n$ swaps between randomly chosen elements in the chosen row.

\paragraph{Minimum change.}
    This mutation uniformly at random chooses a row and two elements in that row. 
    It then increments the value of the first element and decrements the value of the second element.
    If the values cannot be increased/decreased, no changes are made.
    
\paragraph{Ruin and Recreate.}
    This mutation produces a new deterministic random matrix.
    Specifically, it assigns 1 to a randomly selected element in each row and 0 to the rest of the elements.
    
\paragraph{Void.}
    This mutation does not make any changes to the matrix.
    It was introduced to allow changes that affect only one matrix.

\bigskip

In this research, we discretised the elements of the transition matrices. 
Each element could only take values $0, 1/|\mathcal{H}|, 2/|\mathcal{H}|, \ldots, 1$.
Each row needs to add up to 1.
This restriction is upheld by the random generation and optimisation algorithms developed for these matrices. 
Due to time constraints, the use of packages for parameter optimisation such as irace~\cite{Lopez} could not be explored, and thus no evaluation of the effectiveness of the developed matrix optimisation algorithm has been done.

\section{Case study: The three-index assignment problem (AP3)}
\label{sec:ap3}

The three-index assignment problem (AP3) was first mentioned in~\cite{Pierskalla}.
It is an NP-hard combinatorial optimisation problem with many applications.

It is described in~\cite{Fugenschuh2006} as follows: 
``Given $n \in \mathbb{N}$, three sets $I = J = K = \{ 1, \ldots, n \}$, and associated costs $c(i, j, k) \in \mathbb{R}$ for all ordered triples $(i, j, k) \in I \times J \times K$. 
A feasible solution for the AP3 is a set of $n$ triples, such that each pair of triples $(i_1, j_1, k_1)$, $(i_2, j_2, k_2)$ has different entries in every component, i.e., $i_1 \neq i_2$, $j_1 \neq j_2$, and $k_1 \ne k_2$. 
The aim of the AP3 is to find a feasible solution $S$ with minimal costs $\sum_{(i,j,k) \in S} c(i, j, k)$.''

We used three families of AP3 instances for training, validation and testing of each CMCS strategy, all sourced from~\cite{Karapetyan2011}:
\begin{itemize}
    \item
    \textbf{Random instances:} each cost $c(i, j, k)$ is assigned a uniformly distributed random weight from $\{1, 2, \ldots, 100\}$.

    \item
    \textbf{Clique instances:} at first, produce a complete tri-partite graph with partites $I$, $J$ and $K$ and random edge weights in $\{1, 2, \ldots, 100\}$. 
    Then $c(i, j, k)$ is the sum of the weights of the edges $(i, j)$, $(j, k)$ and $(i, k)$.

    \item
    \textbf{Square Root instances:} at first, produce a complete tri-partite graph with partites $I$, $J$ and $K$ and random edge weights in $\{1, 2, \ldots, 100\}$. 
    Then $c(i, j, k)$ is the square root of the sum of squares of the weights of the edges $(i, j)$, $(j, k)$ and $(i, k)$.
\end{itemize}

The training dataset consisted of 12 newly generated instances: 4 instances of size 40 of each type.
All 12 runs were executed in parallel to utilise 12 logical cores of the machine.
The validation dataset consisted of 90 newly generated instances: 10 instances of each type and size (40, 70 or 100). 

The test dataset was used to compare configured algorithms and is the same as that of \cite{Karapetyan2011} and can be found at \url{http://www.cs.nott.ac.uk/~pszdk/?page=publications&key=Karapetyan2011b}. 
It includes 10 instances of each size (40, 70, and 100) and instance type (Random, Clique, and Square Root).

\subsection{Data structures}

This section details the data structures used to hold the information of the problem instances and solutions.

The cost function $c(i, j, k)$ is represented by a three-dimensional integer array of size $n \times n \times n$.

A solution is represented by an array of size $n$ of tuples of size 2; the index of the tuple represents $i$, whereas the values in the tuple are $j$ and $k$.
At all times, the solution is maintained feasible, i.e. $j_1 \neq j_2$ and $k_1 \neq k_2$ for every pair of tuples $(j_1, k_1), (j_2, k_2)$.

While we store solutions as arrays of 2-tuples, we will represent them as sets of 3-tuples in the remainder of the paper.



\section{AP3 components}
\label{sec:ap3-components}

This section describes the components that make up the component pool used as input to the CMCS configurator. 
These components are algorithms that manipulate an AP3 solution.
Some of them were sourced from the AP3 and CMCS literature, while others are newly designed.

\subsection{Neighbourhoods}

To describe the AP3 mutations and hill climbers, it is convenient to first introduce a few neighbourhoods.

\paragraph{Swap:} the neighbourhood includes all the solutions that can be obtained from $S$ by selecting $(i_1, j_1, k_1) \neq (i_2, j_2, k_2) \in S$ and swapping two values: $i_1$ and $i_2$, or $j_1$ and $j_2$, or $k_1$ and $k_2$. 

\paragraph{Shuffle:} the neighbourhood includes all the solutions that can be obtained from $S$ by selecting $(i_1, j_1, k_1) \neq (i_2, j_2, k_2) \neq (i_3, j_3, k_3) \in S$ and shuffling $i_1$, $i_2$ and $i_3$, or $j_1$, $j_2$ and $j_3$, or $k_1$, $k_2$ and $k_3$.

\paragraph{Hungarian:} the neighbourhood includes all the solutions that can be obtained by applying a permutation to the values in one of the three dimensions.
For example, given a solution $(i_1, j_1, k_1)$, $(i_2, j_2, k_2)$, \ldots, $(i_n, j_n, k_n)$, a permutation $\pi$ applied to the first dimension will produce a solution $(\pi(i_1), j_1, k_1)$, $(\pi(i_2), j_2, k_2)$, \ldots, $(\pi(i_n), j_n, k_n)$.

\subsection{Mutations}
Here we discuss AP3 mutations, i.e., components that make random changes to the solution, usually applied to escape a local minimum by worsening its quality.

    \paragraph{Random Swap:} randomly selects a solution from the Swap neighbourhood~\cite{Karapetyan2017}.

    \paragraph{Shuffle Three:} new mutation that randomly selects a solution from the Shuffle neighbourhood.

    \paragraph{Worst Swap:} new mutation that selects the worst solution from the Swap neighbourhood, or returns the original solution if the current solution is a worst one in the Swap neighbourhood.

    \paragraph{First Worsen:}
    is a new mutation that is similar to Worst Swap except that it accepts the first worsening move.

Observe that Random Swap and Shuffle Three mutations are standard stochastic mutations whereas Worst Swap and First Worsen systematically explore a neighbourhood and choose the worst or first worsening move, respectively.

\subsection{Hill Climbers}
The following components are AP3 hill climbers, i.e., components that attempt to improve the solution.

    \paragraph{First Swap:} is a new hill climber that explores the Swap neighbourhood and returns the first improving solution, or returns the original solution if no better solution is found.
    
    \paragraph{Best Swap:} explores the entire Swap neighbourhood and returns the best solution in it~\cite{Karapetyan2017}.

    \paragraph{Hungarian$(d)$:} explores the Hungarian neighbourhood for dimension $d$ and returns the best solution in it.
    Note that the exploration can be reduced to the Assignment Problem and solved in polynomial time by the Hungarian method~\cite{Gabrovsek}.
    Also note that the size of the neighbourhood is exponential.
    Thus, the Hungarian$(d)$ hill climber is a so-called very large neighbourhood search method.
    We used the implementation of the Hungarian method from~\cite{Bhojasia}.

    \paragraph{Min-Dimension Hungarian:} explores the Hungarian neighbourhood for all three dimensions and returns the best solution found~\cite{Gabrovsek}.
    
    \paragraph{All-Dimension Hungarian:} applies Hungarian(1), then Hungarian(2), then Hungarian(3), and then repeats this process until no further improvements can be found~\cite{Gabrovsek}.

    \paragraph{Random Dimension Hungarian:} a new hill climber that chooses the value of $d \in \{1, 2, 3\}$ randomly and applies Hungarian$(d)$.

\section{Experimental Evaluation}
\label{sec:evaluation}

The algorithms for this study were implemented in Java and executed on Acer Predator Triton (6 cores and 12 logical processors) with an Intel Core i7-10750H 2.6GHz processor with 16 GB RAM under Windows 11.
At most six experiments were run in parallel during evaluation (one per physical core).
CMCS and components, however, did not use concurrency.

\subsection{CMCS generation parameters}




The time budget allocated for solving an instance while training was set to 1000~ms. 
According to our experiments, this time was sufficient to model the long-term behaviour of the CMCS configurations.
The time allocated for optimising the transition matrices of each CMCS configuration was 4 minutes for Strategies~A and~B and 2~minutes for each training stage of Strategy~C\@. 

\begin{table}[tb]
\centering
\caption{Generated configurations.}
\label{tab:my-table}
\begin{tabular}{l@{\quad}r@{\quad}r@{\quad}r}
\toprule
Strategy & $|\mathcal{H}|$ & \# component sets & Generation time, min \\
\midrule
A & 2 & 12   & 48     \\
A & 3 & 54   & 216    \\
B & 2 & 12   & 48     \\
B & 3 & 54   & 216    \\
C & 2 & 132  & 792    \\
C & 3 & 2\,862 & 17\,172 \\
\bottomrule
\end{tabular}
\end{table}

Table~\ref{tab:my-table} shows the list of configurations that we generated in this study.
One can see that the generation of Stategy~C configurations is significantly more expensive compared to Strategies~A and~B\@.
Also, the number of configuration components has a great effect on the generation time.
Nevertheless, the generation of a configuration in most of the experiments took no more than a few hours which is well within the time budget for most of the applications and significantly faster than the standard human-based design process.

\subsection{Evaluation results}









\begin{figure}[htb]
\begin{tikzpicture}
\begin{semilogxaxis}[
    compat=newest,
    width=\textwidth,
    height=10cm,
    legend pos=north east,
    xlabel={Time, sec},
    ylabel={Solution error, \%},
    title={},
    grid=major,
    legend cell align=left
]
    \addplot+[red, solid, very thick, mark=*, mark options={fill=red}] table[
        col sep=tab,
	x=Time,
	y=A-2,
    ] {cmcs-2-new.txt};
    \addlegendentry{Strategy A (2-component)}

    \addplot+[blue, solid, very thick, mark=square*, mark options={fill=blue}] table[
        col sep=tab,
	x=Time,
        y=B-2
    ] {cmcs-2-new.txt};
    \addlegendentry{Strategy B (2-component)}

    \addplot+[green, solid, very thick, mark=triangle] table[
        col sep=tab,
	x=Time,
        y=C-2,
    ] {cmcs-2-new.txt};
    \addlegendentry{Strategy C (2-component)}

    \addplot+[red, dashed, thick, mark=*, mark options={fill=red}] table[
        col sep=tab,
	x=Time,
	y=A-3,
    ] {cmcs-3-new.txt};
    \addlegendentry{Strategy A (3-component)}

    \addplot+[blue, dashed, thick, mark=square*, mark options={fill=blue}] table[
        col sep=tab,
	x=Time,
        y=B-3
    ] {cmcs-3-new.txt};
    \addlegendentry{Strategy B (3-component)}

    \addplot+[green, dashed, thick, mark=triangle] table[
        col sep=tab,
	x=Time,
        y=C-3,
    ] {cmcs-3-new.txt};
    \addlegendentry{Strategy C (3-component)}    
\end{semilogxaxis}		
\end{tikzpicture}

\caption{Evaluation of the generated configurations.  The graph shows how the solution error changes throughout the run of CMCS\@.  The solution error is averaged across all the instances in the test set.}
\label{fig:evaluation}
\end{figure}
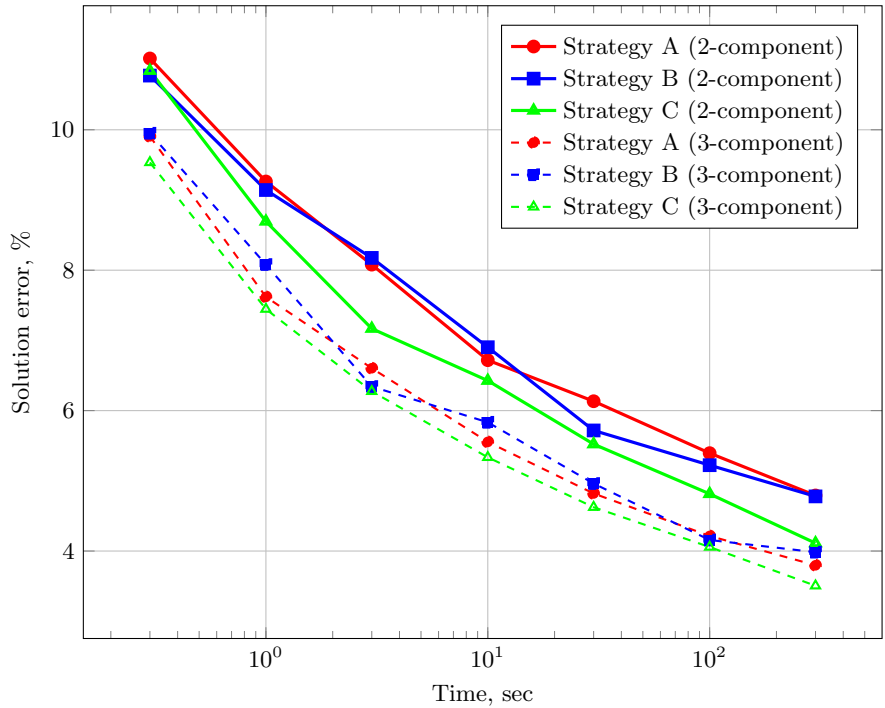

The performance of the obtained configurations is compared in Figure~\ref{fig:evaluation}.
The solid lines correspond to the 2-component configurations whereas the dashed lines correspond to the 3-component configurations.

Among the two-component configurations, Strategy~C is a clear winner for every time budget.
Strategy~B performs similarly to Strategy~A\@.

As expected, moving from 2-component to 3-component configuration improved the solution quality for each strategy.
Otherwise, the results are similar for the 3-component configurations: Strategy~C is a clear winner while Strategies~A and~B perform similarly.

The similarity in performance of Strategies~A and~B most likely indicates a bad choice of the VND stage.
Indeed, our configuration method does not optimise the selection and the sequence of components in VND\@.
Possibly, the algorithm spends too much or too little time on the VND phase.

Nevertheless, the success of Strategy~C proves that CMCS benefits from the addition of an exploitation stage.
This is despite the fact that the generation time for each sub-configuration in Strategy~C was twice smaller than the training time of configurations in Strategies~A and~B\@.

Figure~\ref{fig:cmcs} shows the 3-component Strategy~C configuration generated by our method.
Both sub-configurations have the same component subsets but the transitions are different: sub-configuration~1 uses the mutation more often, which matches our expectations that it would prioritise exploration.
However, there are a few signs that the configuration generation process is not sufficiently robust.
For example, sub-configuration~1 includes a transition from All-Dimension Hungarian to Random-Dimension Hungarian even though All-Dimension Hungarian is guaranteed to find a local minimum in the Hungarian neighbourhood, hence making a subsequent application of the Random-Dimension Hungarian hill-climber useless.
A better configuration optimisation process would eliminate such a transition.

\tikzset{vertex/.style={circle, draw, thick, minimum size=7em}}
\tikzset{edge base/.style={->, >=stealth'}}
\tikzset{improved/.style={edge base, blue!#1!white, bend left=20}}
\tikzset{unimproved/.style={edge base, red!#1!white, bend left=20}}
\tikzset{loop improved/.style={edge base, blue!#1!white, loop above, in=75, out=105, looseness=5}}
\tikzset{loop unimproved/.style={edge base, red!#1!white, loop above, in=60, out=120, looseness=5}}

\begin{figure}[htb]
    \centering
    %
    %
    \begin{subfigure}[t]{0.38\textwidth}
    \footnotesize
    \begin{tikzpicture}
        \useasboundingbox (-2.3,-4.1) rectangle (2.3,2.3);	

    
        \node[vertex, align=center] (RDH) at (-1.5, 0.0) {Random- \\ Dimension \\ Hungarian};
        \node[vertex, align=center] (FW) at (1.5, 0.0) {First \\ Worsen};
        \node[vertex, align=center] (ADH) at (0.0, -2.7) {All- \\ Dimension \\ Hungarian};
        \path (RDH) edge[loop improved=100, line width=1.23] (RDH);
        \path (RDH) edge[improved=100, line width=2.30] (ADH);
        \path (ADH) edge[improved=100, line width=1.47] (FW);
        \path (RDH) edge[unimproved=100, line width=1.48] (FW);
        \path (FW) edge[unimproved=100, line width=2.78] (RDH);
        \path (ADH) edge[unimproved=100, line width=0.74] (RDH);
    \end{tikzpicture}
    \caption{Sub-configuration 1}
    \label{fig:conf-1}
    \end{subfigure}
    \qquad\qquad
    \begin{subfigure}[t]{0.38\textwidth}
    \footnotesize
    \begin{tikzpicture}
        \useasboundingbox (-2.3,-4.1) rectangle (2.3,2.3);	

    
        \node[vertex, align=center] (RDH) at (-1.5, 0.0) {Random- \\ Dimension \\ Hungarian};
        \node[vertex, align=center] (FW) at (1.5, 0.0) {First \\ Worsen};
        \node[vertex, align=center] (ADH) at (0.0, -2.7) {All- \\ Dimension \\ Hungarian};
        \path (RDH) edge[loop improved=100, line width=3.97] (RDH);
        \path (ADH) edge[loop improved=100, line width=1.47, in=165, out=195] (ADH);
        \path (RDH) edge[loop unimproved=100, line width=0.73] (RDH);
        \path (RDH) edge[unimproved=100, line width=1.22] (ADH);
        \path (FW) edge[unimproved=100, line width=1.21] (RDH);
        \path (FW) edge[loop unimproved=100, line width=0.62] (FW);
        \path (ADH) edge[unimproved=100, line width=1.22] (FW);
    \end{tikzpicture}
    \caption{Sub-configuration 2}
    \label{fig:conf-2}
    \end{subfigure}

\caption{
    Generated 3-component Strategy~C configuration (the best configuration found in this study).
    The blue arcs correspond to transitions following improvement of the solution, while the red arcs correspond to the transitions following the solution not being improved).
    The thickness of each arc is proportionate to the frequency of the corresponding transition. 
}
\label{fig:cmcs}
\end{figure}
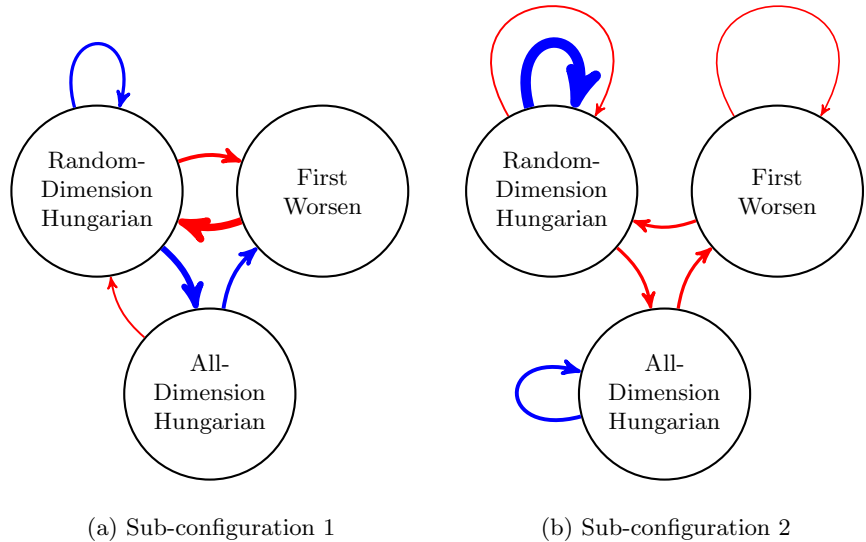

\section{Conclusions}
\label{sec:conclusions}

In this paper, we studied several approaches to intensify exploitation in CMCS\@.
Specifically, we designed two new CMCS strategies and compared them to the original strategy (Strategy~A)\@. 
Our experiments confirm that CMCS benefits from an exploitation mechanism; Strategy~C has clearly outperformed Strategy~A\@.
(Strategy~B demonstrated performance similar to that of Strategy~A -- possibly indicating some ineffective design choices.)
On the other hand, generation of a Strategy~C configuration is computationally expensive compared to Strategies~A and~B\@.

In future, we would like to improve the generation algorithms to study the performance of larger Strategy~C configurations and compete with the more powerful AP3 metaheuristics from the literature.
We would also like to test our new CMCS strategies on other combinatorial optimisation problems to confirm our findings and assess the effects of various parameters introduced in Strategies~B and~C: the ordering of components within VND in Strategy~B, the time allocated to sub-configuration~1 in Strategy~C, etc.




%
%
%
\bibliographystyle{splncs04}
\bibliography{refs}

\end{document}